\documentclass[sigconf]{acmart}
\usepackage{xcolor}
\usepackage{algorithm}
\usepackage{algpseudocode}
\usepackage{tcolorbox}

\AtBeginDocument{%
  \providecommand\BibTeX{{%
    \normalfont B\kern-0.5em{\scshape i\kern-0.25em b}\kern-0.8em\TeX}}}

\setcopyright{acmcopyright}
\copyrightyear{2018}
\acmYear{2018}
\acmDOI{XXXXXXX.XXXXXXX}

\acmConference[Conference acronym 'XX]{Make sure to enter the correct
  conference title from your rights confirmation emai}{June 03--05,
  2018}{Woodstock, NY}

\acmPrice{15.00}
\acmISBN{978-1-4503-XXXX-X/18/06}

\author{MSVPJ Sathvik}
\affiliation{%
  \institution{IIIT Dharwad}
  \country{India}
}
\email{20bec024@iiitdwd.ac.in}

\author{Surjodeep Sarkar}
\affiliation{%
  \institution{UMBC }
  \city{Baltimore, MD}
  \country{USA}
}
\email{ssarkar1@umbc.edu}

\author{Chandni Saxena}
\affiliation{%
 \institution{The Chinese University of Hong Kong}
 \country{Hong Kong}}
 \email{csaxena@cse.cuhk.edu.hk}

\author{Sunghwan Sohn}
\affiliation{%
  \institution{Mayo Clinic}
  \state{Rochester, MN}
  \country{USA}}
\email{sohn.sunghwan@mayo.edu}

\author{Muskan Garg}
\affiliation{%
  \institution{Mayo Clinic}
  \state{Rochester, MN}
  \country{USA}}
  \email{garg.muskan@mayo.edu}

\begin{document}

\title{InterPrompt: Interpretable Prompting for Interrelated Interpersonal Risk Factors in Reddit Posts}

\renewcommand{\shortauthors}{Sathvik et  al.}

\begin{abstract}
Mental health professionals and clinicians have observed the upsurge of mental disorders due to Interpersonal Risk Factors (IRFs). To simulate the human-in-the-loop triaging scenario for early detection of mental health disorders, we recognized textual indications to ascertain these IRFs: \textit{Thwarted Belongingness} (\textit{TBe}) and \textit{Perceived Burdensomeness} (\textit{PBu}) within personal narratives. In light of this, we use N-shot learning with GPT-3 model on the IRF dataset, and underscored the importance of fine-tuning GPT-3 model to incorporate the context-specific sensitivity and the interconnectedness of textual cues that represent both IRFs. 

In this paper, we introduce an \textit{Interpretable Prompting (InterPrompt)} method to boost the attention mechanism by fine-tuning the GPT-3 model. This allows a more sophisticated level of language modification by adjusting the pre-trained weights. Our model learns to detect usual patterns and underlying connections across both the IRFs, which leads to better system-level explainability and trustworthiness. The results of our research demonstrate that all four variants of GPT-3 model, when fine-tuned with InterPrompt, perform considerably better as compared to the baseline methods, both in terms of \textit{classification} and \textit{explanation generation}.

\end{abstract}



\maketitle
\section{Background}
Every year, over 700,000 lives are lost to suicide, as reported by the World Health Organization. The number of attempted suicides stemming from each completed suicide surpasses this figure,\footnote{\url{https://www.who.int/news-room/fact-sheets/detail/suicide}} emphasizing the significant role that \textit{interpersonal relationships} play in an individual's life. The Interpersonal-Psychological Theory of Suicide (IPTS)~\cite{joiner2009interpersonal}, asserts the severity of suicidal tendencies as the result of three interconnected factors: (i) \textit{Acquired Capability} for committing suicide, (ii) \textit{Thwarted Belongingness} (\textsc{TBe}) characterized by feelings of social isolation and disconnectedness, and (iii) \textit{Perceived Burdensomeness} (\textsc{PBu}) involving a sense of being a burden on society. Consequently, \textsc{TBe} and \textsc{PBu} together act as Interpersonal Risk Factors (IRFs), signifying adequate immediate causes of suicide. Social media platforms, especially those offering anonymity (e.g., subreddit in Reddit), are increasingly becoming a refuge for individuals whose psychological stigma deters them from seeking professional mental healthcare~\cite{tay2018mental}. These platforms facilitate easy and anonymous sharing of personal experiences, creating a secure, empathetic environment for self-expression~\cite{roy2023proknow}. The growing demand for quality mental healthcare, coupled with a scarcity of clinical psychologists, highlights the importance of computational models in analyzing mental health through social media posts~\cite{torous2021growing}.

\begin{figure}
    \centering
    \includegraphics[width=0.40\textwidth]{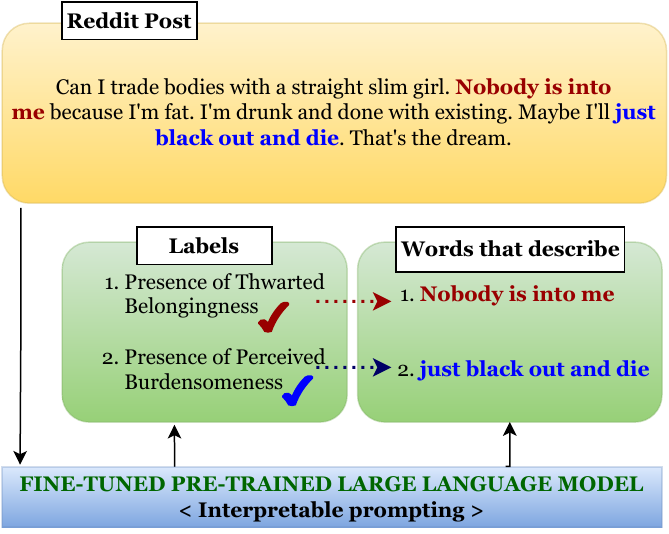}
    \caption{Schema of the proposed work for identifying textual cues while determining Interpersonal Risk Factors in Reddit Post.}
    \label{fig:1}
\end{figure}

To address the above issues, we draw inspiration from the notion of \textit{Safety, Explainability, and Trustworthiness} advocated by the social NLP community~\cite{roy2023proknow}, that emphasizes \textit{``interpretability''} and \textit{``system-level explainability''} in computational models for mental health analysis. In this paper, we define ``Explainability'' as the linguistic cues that shed light on the presence of \textsc{TBe} and \textsc{PBu} within text. 
We take into consideration of GPT-3's capability to generate sentiment-driven text with a human-like quality, leading us to regard it as an exceptionally fitting model for mental health analysis~\cite{brown2020language,xie2022pre,yang2023evaluations}. However, in order to build more comprehensive and explainable model that can effectively understand the linguistic cues, we introduce ``N-shot learning''~\cite{weifinetuned} followed by our \textbf{``Interpretable Prompting (InterPrompt)''} approach for task-specific fine-tuning of the GPT-3 model. 
As such, we aim to enhance the performance of GPT-3 model and tailor it to a downstream task of explainable identification of IRFs, while still benefiting from the knowledge and understanding, the model gained during the initial pre-training phase (see Figure~\ref{fig:1}).

% \textbf{Motivation: @ MUSKAN MAM} The mental health of a person can be monitored by his social media posts. The explanations for the classification would better help the psychologist and users by transparency and trustworthiness.

\begin{figure*}[t]
    \centering
    \includegraphics[width=0.96\textwidth]{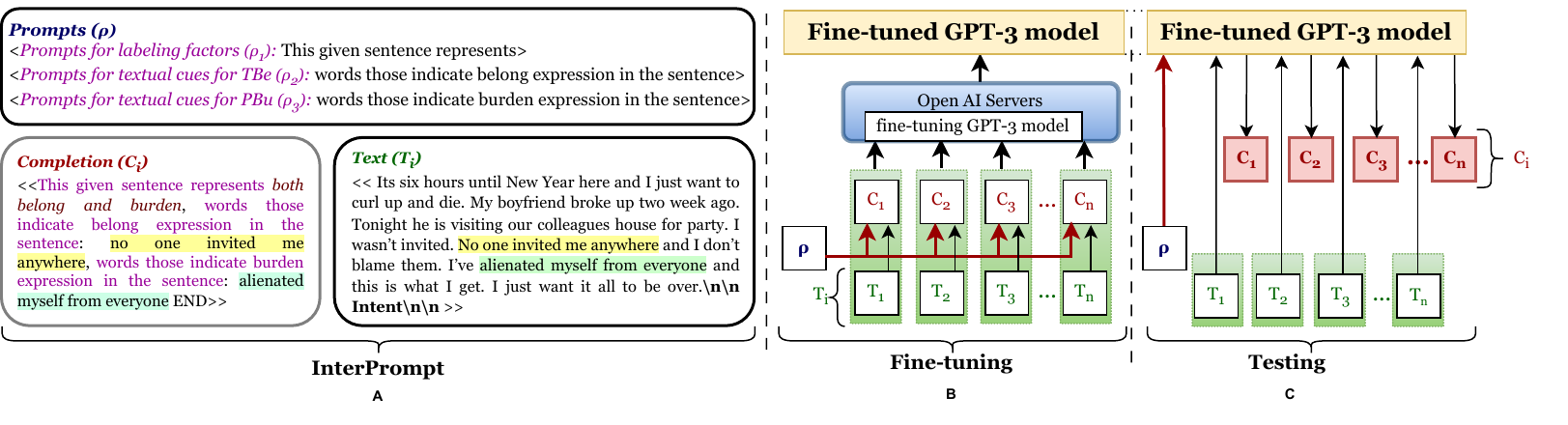}
    \caption{Overview of the InterPrompt-driven fine-tuning of GPT-3 model. \textsc{TBe}: Thwarted Belongingness, and \textsc{PBu}: Perceived Burdensomeness.}
    \label{fig2cikm}
\end{figure*}

We present our contributions in three-fold ways: (i) we provide an in-depth analysis that demonstrates the inefficiency of GPT-3 model, when employing ``N-shot learning'' for task-specific mental health analysis, (ii) we introduce \textit{InterPrompt}, which effectively incorporates task-specific context to enhance the \textit{interpretability} and \textit{System-level explainabilty} of the model, and (iii) we compare the InterPrompt-driven fine-tuned GPT-3 model with baselines to provide empirical evidence of the advantages and advancements offered by our approach.
To the best of our knowledge, we are the first to design prompt to comprehend interrelated IRFs in a given text.

% \section{Related Work}

\section{Proposed Method}

\paragraph{\textbf{Problem Definition}}

%The maximum length of each Reddit post is 300 words that contains multiple sentences. 
Let $\mathcal{D}$ be the given IRF dataset. Given a text $T= \{S_1, S_2, S_3,..., S_N,$ $ S_{sep}\}$ which is a sequence of $N$ sentences, where $S_{sep} = [SEP]$ represents
``\verb|\n\nIntent:\n\n|'' as illustrated in Figure \ref{fig2cikm}. In our problem formulation, we first define the \textit{textual cues} for \textbf{TBe} in the given text $T$ which is denoted as $\mu_i(\epsilon)$, where $\mu_i$ represents \textit{textual cues} and $\epsilon$ represents the \textbf{TBe}. We define the \textit{label} for \textbf{TBe} as ($y_i(\epsilon)$), where $y_i$ represents the label for a given factor.
Secondly, we define the \textit{textual cues} for \textbf{PBu} in the given text $T$ which is denoted as ($\mu_i(\upsilon)$), where $\mu_i$ represents \textit{textual cues} and $\upsilon$ represents the \textbf{PBu}. We define the \textit{label} for \textbf{PBu} as ($y_i(\epsilon)$), where $y_i$ represents the label for a given task. We initially transform the factor labels from their binary representation into text format. We further classify our approach in three phases: (A) InterPrompt, (B) fine-tuning GPT-3 model, and (C) testing phase as illustrated in Figure \ref{fig2cikm}. 

For \textbf{InterPrompt}, we define notations for text, prompts, and completion. We consider prompts denoted as $\rho = \{\rho_1, \rho_2, \rho_3,..., \rho_m\}$ where $m=3$ and we define  $\rho=\{\rho_1, \rho_2, \rho_3\}$ as a set of (i) \textit{labeling factors} ($\rho_1$): \texttt{This given sentence represents \{belong, burden, both belong and burden, neither belong nor burden\}}, (ii) \textit{textual cues for \textsc{TBe}} ($\rho_2$): \texttt{`word those indicate belong expression in the sentence...'}, (iii) \textit{textual cues for \textsc{PBu}} ($\rho_3$): \texttt{`word those indicate burden expression in the sentence...'}. The idea of InterPrompt is to generate the \textit{story completion} ($C_i$) the process of generating or completing text ($T_i$) based on the given prompts ($\rho$) 
%denoted as $C_i$ as a function of text ($T_i$), prompts ($\rho$). %, factor labels ($y_i$), and textual cues ($\mu_i$).% 
as shown in Equation~\ref{eqn1}.
% \begin{equation}
% \label{eqn1}
%     C_i = f(\rho \cdot T_i) %y_i, \mu_i)|T_i)
% \end{equation}
\begin{equation}
\label{eqn1}
    C_i = f(\rho_1 + y_i(\epsilon, \upsilon))) + f(\rho_2 + \mu_i(\epsilon)) + f(\rho_3 + \mu_i(\upsilon)) % \cdot T_i) %y_i, \mu_i)|T_i)
\end{equation}
Here, $f(.)$ is a function that generates multi-task story completion using the three prompts. $\rho_1$ is used for labeling factors, $\rho_2$ is used for textual cues for \textsc{TBe}, and $\rho_3$ is used for textual cues for \textsc{PBu}. 
% Thus, the completion is defined as
% \begin{equation}
% \label{eqn2}
%     C = \sum_{i=1}^{k} C_i; \quad k \subseteq N
% \end{equation}
% where N is the number of samples in a given corpus. 

\textbf{For fine-tuning GPT-3 model}, we construe the story completion to fine-tune GPT-3 model ($G$) using OpenAI servers. GPT-3 model finds the probability of obtaining required output for the given input. We define the probabilities $P=\{P_1, P_2, P_3\}$ for three tasks of identifying $\hat{y_i}(\epsilon, \upsilon), \hat{\mu_i}(\epsilon), \hat{\mu_i}(\upsilon)$, respectively. We train our model with $k$ samples:

\begin{equation}
\begin{aligned}
    \hat{y_i}(\epsilon, \upsilon)=P_1(y_i(\epsilon, \upsilon)| T_i, \rho_1); \quad i \in k,\\
    \hat{\mu_i}(\epsilon) = P_2(\mu_i(\epsilon)| T_i, \rho_2); \quad i \in k,\\ 
    \hat{\mu_i}(\upsilon) = P_3(\mu_i(\upsilon)| T_i, \rho_3); \quad i \in k,\\ 
\end{aligned}
\end{equation}

where $k$ is the number of samples used to fine-tune different variants of GPT-3 model using OpenAI servers. The story completion is a prompt designed to capture multi-tasking of the model as:
\begin{equation}
\label{eqn4}
    C_i= \rho_1 + \hat{y_i}(\epsilon, \upsilon) + \rho_2 + \hat{\mu_i}(\epsilon) + \rho_3 + \hat{\mu_i}(\upsilon)
\end{equation}

\paragraph{\textit{Loss Function for fine-tuning GPT-3}}: To achieve story completion by taking into account multiple prompts and focusing on a combined loss function that incorporates both the structured extraction of entities and the generation of coherent text. We define the combined loss for our task as:

\begin{equation}
L_{combined} = \lambda_1 \cdot L_1 + \lambda_2 \cdot L_2 + \lambda_3 \cdot L_3
\end{equation}

where $L\_1, L\_2, L\_3$ are defined as loss functions for predicting $\hat{y_i}(\epsilon, \upsilon), $ $\hat{\mu_i}(\epsilon), \hat{\mu_i}(\upsilon)$, respectively. For a given sequence $S_i$ of text $T_i$ and the number of generated tokens as $E_i$, we define loss functions as:
\begin{equation}
\begin{aligned}
    L_1 (y_i, \hat{y_i}) = \left(-\sum_{t=1}^{S_i} \sum_{i=1}^{E_i} y_{t,i} \log(\hat{y}_{t,i}|\rho_1)\right)\\
    L_2 (\mu_i (\epsilon), \hat{\mu_i}(\epsilon)) = \left(-\sum_{t=1}^{S_i} \log P(\hat{\mu_i}(\epsilon) | \rho_2)\right)\\
    L_3 (\mu_i (\upsilon), \hat{\mu_i}(\upsilon)) = \left(-\sum_{t=1}^{S_i} \log P(\hat{\mu_i}(\upsilon) | \rho_3)\right)
\end{aligned}
\end{equation}

Here, $L_1$ is a token level cross-entropy loss functions for entity extraction; $L_2$ and $L_3$ are the maximum likelihood estimation loss for text generation. $\lambda_1, \lambda_2$ and $\lambda_3$ are the weights that balance the contribution of all three loss components for identifying explainable IRFs during fine-tuning GPT-3 model.
We define the process of fine-tuning $G$ on OpenAI servers as a function ($O(G)$). The model $G$ is fine-tuned by optimizing a loss function $L$ based on the output of the GPT-3 model on the story completion $C_i$ and the true labels $C_i$. The equation can be represented in a simplified form as:
\begin{equation}
\label{eqn2}
   O(G) = \arg\min_{k}(L_{combined}(C_i, \hat{C_i}))
\end{equation}

where our GPT model is fine-tuned by adjusting the pre-trained weights to minimize loss function.
% \begin{equation}
% \label{eqn2}
%     Cf_i = f_f (C_i \cdot (y_i, \mu_i))
% \end{equation}
% We fine-tune the GPT-3 model with \textbf{prompts} and \textbf{completion} as significant inputs (see Algorithm~\ref{alg:GPT}).

% Given a text 
% $T= \{S_1, S_2, S_3,..., S_N,$ $ S_{sep}\}$ which is a sequence of $N$ sentences, where $S_{sep} = [SEP]$ represents
% %We add the suffix [SEP] in the end of the text as
% ``\verb|\n\nIntent:\n\n|'' as illustrated in Figure \ref{figicot}. We first denote the \textit{textual cues} for \textbf{TBe} in the given text $T$ which is denoted as $\mu_i(\epsilon)$, where $\mu_i$ represents \textit{textual cues} and $\epsilon$ represents the \textbf{TBe} from the . We denote the \textit{label} for \textbf{TBe} as ($y_i(\epsilon)$), where $y_i$ represents the label for a given task.

 % the \textit{textual cues} for \textbf{PBu} ($\mu_i(\upsilon)$) suggesting the \textit{label} for \textbf{PBu} ($y_i(\upsilon)$). 

% The sentences $S_i$ within the given text $T$ containing textual cues $C_i$ and is labeled as $y_i$ for IRF: \textsc{TBe} ($\epsilon$) and \textsc{PBu} ($\upsilon$). 

\textbf{For testing phase}, %the \textbf{InterPrompt-driven fine-tuned GPT-3 model (fine-tuned GPT-3)} 
we label a given text ($T_i$) combined with input $\rho$ as $\sum_{i=1}^{k} (C_i|T_i, \rho);$ where $k \subseteq N$. 
% \begin{equation}
%      Ct_i= f_t(\hat{y_i}(\epsilon,\upsilon), \mu(\epsilon,\upsilon)  | (T_i, \rho))
% \end{equation}
We decode the single story completion $\hat{C_i}$ to find labels ($y_i$) and textual cues ($\mu$) for both TBe ($\epsilon$) and PBu ($\upsilon$). The textual cues is a single most expressive continuous expression of words obtained against $\mu$. Labels ($y_i$) are mapped back from string to binary labels.

\section{Experiments}

% This section delineate the use of InterPrompt to fine-tune the GPT-3 model used in this experiments.

\paragraph{\textbf{Dataset}} Central to the IPTS, we emphasize the exploration of textual cues related to interconnected interpersonal risk factors in a given text. This approach upholds our fine-grained observations, robust arguments and are reinforced by statistical analysis of IRF dataset~\cite{garg2023annotated}. %The dataset contains five columns representing [Text, Label (\textsc{TBe}), Textual Cues (\textsc{TBe}), Label (\textsc{PBu}), Textual Cues (\textsc{PBu})]. 
We observe that the ratio of instances where PBu equals 1 to instances where TBe equals 1 is approximately 32\% higher than the increase in the ratio of PBu equals 1 to TBe equals 0, as illustrated in Table~\ref{tab1}. Our finding emphasizes a stronger correlation between TBe and PBu.

%We notice the increase in the ratio of $\textsc{PBu}=1$ to $\textsc{TBe}=1$ is $\approx 32\%$ more than increase in the ratio of $\textsc{PBu}=1$ to $\textsc{TBe}=0$ , highlighting a stronger correlation between \textsc{TBe} and \textsc{PBu}. The preview of the dataset is illustrated in Table~\ref{tab1}. 

\begin{table}[t]
\small
    \centering
    \begin{tabular}{c|c|c}
       \toprule[1.5pt]
       & \textsc{\textsc{PBu}: 0} & \textsc{\textsc{PBu}: 1}\\
        \toprule[1.5pt]
        \textsc{\textsc{TBe}: 0} & 1123 & 472 \\
        \midrule
        
        \textsc{\textsc{TBe}: 1} & 1252 & 675 \\
        \midrule
        \%$\Delta$ & 129/1123 $=0.1148$ & 203/472 $=0.4301$
        \\
       \bottomrule[1.5pt]
    \end{tabular}
    \caption{Dataset statistics for Thwarted Belongingness (TBe) and Perceived Burdensomeness (PBu) representing the number of samples in IRF dataset for corresponding labels.}
    \label{tab1}
\end{table}

\begin{table}[!t]
     \small
    \centering
    \caption{Statistical significance test for classifiers. Here, p-val is p-value, IPG is InterPrompt-driven fine-tuned GPT-3 model, A is \texttt{ada} variant, B is \texttt{babbage} variant, C is \texttt{curie} variant, D is \texttt{davinci} variant}
    \label{tab:3}
    \begin{tabular}{c|cc|cc|cc}
        \toprule
                \textbf{Models}&  \multicolumn{2}{c|}{IPG [B]} &  \multicolumn{2}{c|}{IPG [C]} &  \multicolumn{2}{c}{IPG [D]} \\
       \midrule
        & \textbf{t-test}  & \textbf{p-val} &\textbf{t-test} & \textbf{p-val} &\textbf{t-test} & \textbf{p-val}    \\ \midrule
         IPG [A]&0.053&0.957&-0.639&0.523&-0.426&0.670   \\ 
         IPG [B]&-&-& -0.693&0.489& -0.479&0.632  \\ 
         IPG [C]& -&-&-&-&0.213&0.831\\ 
         % Davinci &  -0.4794&0.6317&0.2132&0.8312&-&- \\ 
        % GPT Babbage and GPT Davinci& -0.4794 & 0.6317   \\ 
        % GPT Curie and GPT Davinci& 0.2132 & 0.8312   \\ 
        \bottomrule
    \end{tabular}
\end{table}

\begin{table*}[ht!]
\centering
\small
\caption{Comparison of InterPrompt driven fine-tuned GPT-3 variants  with baseline models over IRF dataset}
\label{tab:results}
\begin{tabular}{l|ccc|c|ccc|c}
 \toprule[1.5pt]
\textbf{Model}      & \multicolumn{4}{c|}{\textsc{\textbf{Thwarted Belongingness}}}& \multicolumn{4}{c}{\textsc{\textbf{Perceived Burdensomeness}}} \\
& \textbf{Precision} & \textbf{Recall} & \textbf{F1-score} & \textbf{Accuracy}& \textbf{Precision} & \textbf{Recall} & \textbf{F1-score} & \textbf{Accuracy}  \\ \midrule[1pt]
\textbf{BERT}    & 69.70  &  76.97    & 72.30 & 68.97 & 56.47     &  53.00 &   52.20   &    72.56     \\
\textbf{RoBERTa}    & 71.23 & 73.54   & 71.35   & 68.97 & 67.27       & 37.52       &   45.51 &  74.93     \\
\textbf{DistilBERT}       & 70.24 & 74.08 & 71.15 & 68.50 & 51.15 & 31.89 & 36.93 &      71.71  \\
\textbf{MentalBERT}    & 77.97 & 77.40 & 76.73 & 75.12   &   64.22    &  65.75      & 62.77   & 78.33    \\
\midrule
\textbf{OpenAI+LR}  & 79.00 &	\textbf{83.59} &	81.23 &	78.62 &	82.66 &	63.08 &	71.55 &	84.58     \\
\textbf{OpenAI+RF}  & 79.06 &	80.68 &	79.86 &	77.48 &	\textbf{83.33} &	49.23 &	61.90 &	81.36    \\
\textbf{OpenAI+SVM} & 81.31 &	80.34 &	\textbf{80.83} &	\textbf{78.90} &	79.15 &	74.77 &	\textbf{76.90} &	\textbf{86.19}  \\
\textbf{OpenAI+MLP} & \textbf{81.40} &	75.56 &	78.37 &	76.92 &	72.08 &	\textbf{77.85} &	74.85 &	83.92    \\
\textbf{OpenAI+XGB} & 81.22 &	79.83 &	80.52 &	78.62 &	80.36 &	68.00 &	73.67 &	85.05    \\
\midrule
\textbf{GPT-3 Zero-shot} & 63.78 &	21.54 &	32.21 &	51.63 &	27.56 &	16.28 &	20.47 &	61.42    \\
\textbf{GPT-3 One-shot}  & 61.15 & 84.57 & 70.98& 63.12   & 34.84 & 86.05 & 49.60 & 46.67    \\
\textbf{GPT-3 Few-shot}  & 57.42 & 94.68 & 71.49 & 59.72   & 32.16 & 98.14 & 48.45 & 36.31    \\
\midrule
\textbf{\textsc{Fine-tuned GPT-3 [A]}} & 86.14 &	83.93 &	85.02 &	83.63 &	76.72 &	79.08 &	77.88 &	86.19    \\
\textbf{\textsc{Fine-tuned GPT-3 [B]}} & \textbf{86.38} &	84.21 &	\textbf{85.12} &	\textbf{84.29} &	\textbf{81.24 }&	80.69 &	80.96 &	87.08    \\
\textbf{\textsc{Fine-tuned GPT-3 [C]}} & 85.17 &	\textbf{87.35} &	86.24 &	84.58 &	79.94 &	80.92 &	80.43 &	\textbf{87.89}    \\
\textbf{\textsc{Fine-tuned GPT-3 [D]}} & 84.59 &	83.01 &	83.80 &	82.83 &	80.12 &	\textbf{84.17} &	\textbf{82.13} &	87.54    \\
 \bottomrule[1.5pt]
\end{tabular}
\end{table*}

\begin{table*}[ht!]
     \small
    \centering
    \caption{Performance evaluation of generated explanations through similarity measures}
    \label{tab:Exp_TB}
    \begin{tabular}{l|cccc|cccc}
        \toprule[1.5pt]

        \textbf{Model Name} & \multicolumn{4}{c|}{\textsc{\textbf{Thwarted Belongingness}}} & \multicolumn{4}{c}{\textsc{\textbf{Perceived Burdensomeness}}} \\
        
         &\textbf{Rouge-1} & \textbf{Rouge-L} &  \textbf{BLEU-1} &\textbf{EM}&\textbf{Rouge-1} & \textbf{Rouge-L} &  \textbf{BLEU-1}& \textbf{EM} \\ \midrule
        \textbf{MentalBERT+LIME} &  0.2202   & 	- & 0.1509 &	- & 0.2425	&	-	&	0.1706& - \\
% & \textsc{PBu} & 0.2425 &	- &	- &	0.1706 & - & 	-	& - \\
% \midrule
        \textbf{MentalBERT+ SHAP} &  0.2415 & - & 	0.1593 & - & 0.2601 &		-&0.1718&	- \\
        % \midrule
        % \textit{MentalBERT+ SHAP} & \textsc{TBe} & 0.6597 & 0.1477&0.6591 &0.6373 &0.300 & 0.5787&0.6442 \\
        % & \textsc{PBu}  &	- &	- &	 &	- &	-	& - \\ 
        \midrule
        \textbf{\textsc{Fine-tuned GPT-3 [A]}}  & 0.6597 & 0.6591 &0.6373 & 0.5787  & 0.7738  & 0.7738 & 0.7556  & 0.6993 \\ 
     
        \textbf{\textsc{Fine-tuned GPT-3 [B]}} & 0.6631 & 0.6628&0.6412  & \textbf{0.5816} & 0.7832  & 0.7831 & 0.7693  & 0.7163  \\ 
        % \midrule
        \textbf{\textsc{Fine-tuned GPT-3 [C]}}  &0.6637  &0.6633 &0.6377  &0.5603 & \textbf{0.7989}  & \textbf{0.7989} & \textbf{0.7846}  & \textbf{0.7348} \\ 
        % \midrule
        \textbf{\textsc{Fine-tuned GPT-3 [D]}} &\textbf{0.6809}  &\textbf{0.6805} &\textbf{0.6532} &0.5801  & 0.7771 &  0.7771 & 0.7579  & 0.6965 \\ \bottomrule[1.5pt]
    \end{tabular}
    
\end{table*}
\paragraph{\textbf{N-Shot Learning}}
% We examine N-shot learning with GPT-3 on IRF dataset~\cite{garg2023annotated} to determine the presence of \textsc{TBe} and \textsc{PBu}. Sensitivity to exemplars is a key consideration of prompting approaches. The permutation of N-shot learning demonstrate the accuracy of GPT-3 classifiers tasks for mental health domain to range around \textit{near chance} (about 40\% to 60\%) (see Table~\ref{tab:results}). We first employ \textit{zero-shot prompting} with GPT-3 \texttt{davinci}, resulting in 51.63\% and 61.42\% accuracy for \textsc{TBe} and \textsc{PBu}, respectively. We further implement \textit{one-shot learning and few-shot learning} approach by setting \textit{one} and \textit{five} initial input examples, respectively. 

Central to the sensitivity to exemplars in prompting approaches, we investigate the permutations of N-shot learning in GPT-3 prompting to evaluate the effectiveness of GPT-3 classifiers in the mental health domain. The results indicate that the accuracy of GPT-3 classifiers for TBe and PBu tasks falls within the range of \textit{near chance}, typically around 40\% to 60\% (see Table~\ref{tab:results}). We first employ the ``zero-shot prompting'' approach yielding an accuracy of 51.63\% for TBe and 61.42\% for PBu. As such, we proceed to implement the ``one-shot learning'' and ``few-shot learning'' approaches, using a single initial input example and eight initial input examples, respectively. The decline in performance observed with N-shot learning can be attributed to the contextual sensitivity, underscoring the importance of task-specific fine-tuning of pre-trained language models (PLMs).

\subsection{Experimental Setup}
% \paragraph{\textbf{Experimental Setup}} 
For our experimental purposes and evaluation, we utilize the IRF, a Reddit dataset comprising 3522 posts which is divided into three sets: training (1972 posts), validation (493 posts), and testing (1057 posts)~\cite{garg2023annotated}. We fine-tune all four variants of GPT-3 using the training dataset at OpenAI server, namely, (i) ada $\rightarrow$ [A], (ii) babbage $\rightarrow$ [B], curie $\rightarrow$ [C], davinci $\rightarrow$ [D]. We keep the parameters at their default settings. We compare our model with baseline models: (i) pre-trained lanaguage models: BERT~\cite{devlin2018bert}, RoBERTa~\cite{liu2019roberta}, DistilBERT~\cite{sanh2019distilbert}, MentalBERT~\cite{ji2021mentalbert}, (ii) OpenAI + machine learning models~\cite{garg2023annotated}.

\paragraph{\textbf{Evaluation Metrics}} 
%For experiments and evaluation, we use the IRF, a Reddit dataset containing 3522 posts (train: 1972, val: 493, test: 1057) (see Table~\ref{tab:2})~\cite{garg2023annotated}.
% In this section, we introduce the details of using four variants of GPT-3 for mental health classification task, including datasets, fine-tuning and evalation metrics.   
%\subsection{GPT} GPT (Generative Pre-trained Transformer) is a LLM model developed by OpenAI\footnote{https://platform.openai.com/docs/model-index-for-researchers} to generate a wide range of written content such as articles, poetry, stories, news reports, and dialogues. %GPT-3~\cite{brown2020language} represents the third iteration of GPT models, capable of executing a wide range of tasks such as question answering, text summarization and parsing, translation, and classification. The model interacts with users through the input of text prompts, generating corresponding text completions based on the input prompt. GPT-3 has shown its impressive few-shot learning ability and human-like text generation. 
% \footnote{see Appendix~\ref{baselines} for more details.} 
% To provide precise information and avoid generating extraneous text with fine-tuning of GPT-3, we attach a suffix to the input as ``\verb|\n\nIntent:\n\n|'' and to the output as ``\verb|END|''. 

The best performing baseline among pre-trained language models, MentalBERT, generate explanations using two explainable NLP methods: LIME~\cite{lime} and SHAP~\cite{shap}. Our fine-tuned GPT-3 models generate explanations along with classified labels. We assess the classification performance of all models using standard metrics such as Precision, Recall, F1-Score, and Accuracy. To evaluate the quality of the generated explanations, we compare them with ground-truth explanations using similarity check algorithms including Recall-Oriented Understudy for Gisting Evaluation (ROUGE)~\cite{lin2004rouge}, BiLingual Evaluation Understudy (BLEU)~\cite{papineni2002bleu}, and Exact Match (EM) scores, comparing them with ground-truth explanations. The goal was to determine the faithfulness and accuracy of the explanations generated by our fine-tuned GPT-3 models.

\subsection{Results and Discussion}
\paragraph{\textbf{Classification Model}}
 %However, N-shot learning techniques decline the performance due to generalized nature of Generative AI.
In our study, we found that our models consistently outperformed all existing baselines as illustrated in Table~\ref{tab:results}. Notably, we observed higher recall values compared to precision, indicating that our models accurately recognized the correct labels. Based on our experiments, "fine-tuned GPT-3 [B]" gives best performance in terms of accuracy, and "fine-tuned GPT-3 [C]" and "fine-tuned GPT-3 [D]" outperforms all the other classifiers for TBe and PBu in terms of F1-score, respectively. %Furthermore, we notice variations in the values of evaluation metrics for different variants, indicating the necessity of conducting significance tests on experimental results.

\paragraph{\textbf{Statistical Significance}}
As we observe minor variation in \textit{accuracy} and \textit{F1-score} among all the variants, we perform the statistical significance test~\cite{t-test,p-value}. We used the student's t-test~\cite{t-test2} to %decide if the resulting values have significant difference %evaluate the importance and reliability of fine-tuning GPT-3 variants for classification tasks by 
%by 
assess the statistical significance with $p-value = 0.05$ (see Table~\ref{tab:3}).
We notice significant differences in the performance of different variants of fine-tuned GPT-3 models, indicating the best performance classifiers with highest values of accuracy.

\paragraph{\textbf{Generated Explanations}}
%A well established model MentalBERT is a contextualized pre-trained model for mental healthcare domain. % and we observe ROUGE-1 scores for \textsc{TBe} and \textsc{PBu} for Local Interpretable Model-Agnostic Explanations (LIME: 22.02\% and 24.25\%) and Shapley Additive Explanations (SHAP: 24.15\% and 26.01\%), respectively. 
%To evaluate the explanations generated by MentalBERT and all the variants of our model, we examine ROUGE scores, BLEU scores, Exact Match score and NDCG score. We observe the scores for similarity among generated text-spans (explanations) and the ground-truth interpretations (human-annotated), and found that 
We further evaluate the explanations generated and compare them with ground-truth explanations to examine increased reliability and robustness in fine-tuned GPT-3 models. 
The explanations generated by LIME and SHAP lack continuity, which restricts their ability to generate ROUGE-L and EM scores. Our observations indicate that the fine-tuned GPT-3 models labeled as "fine-tuned GPT-3 [D]" and "fine-tuned GPT-3 [C]" exhibit the highest performance for TBe and PBu, respectively, surpassing all other methods considered. This finding suggests an enhanced level of reliability and efficacy in these particular models. More specifically, % the obtained values for ROUGE-1 and BLEU metrics demonstrate 
our analysis reveals notable performance by the "fine-tuned GPT-3 [D]" model in the TBe task, achieving a ROUGE-1 score of 68.09\% and a BLEU score of 65.32\%. Similarly, the "fine-tuned GPT-3 [C]" model showcases impressive performance in the PBu task, attaining a ROUGE-1 score of 79.89\% and a BLEU score of 78.46\% (see Table~\ref{tab:Exp_TB}).
Overall, the system-level explainability for PBu demonstrates superior performance compared to TBe, leading to higher F1-scores and Accuracy measures.
\paragraph{\textbf{Ethics and Broader Impact}}

We are dedicated to upholding ethical principles and protecting user privacy~\cite{yoo2023discussing,sawhney2023much}. In accordance with this commitment, we rigorously comply with OpenAI's terms and conditions and refrain from divulging our models or OpenAI key. In order to safeguard user privacy and minimize the risk of misuse, all examples presented in this paper have been anonymized, obfuscated, and paraphrased~\cite{zirikly2019clpsych}. This NLP-focused task finds its practical utility in the preliminary assessment of individuals on social media platforms during in-person sessions of mental health triaging, and clinical diagnostic interviewing~\cite{artificialintelligencenewsBabylonHealth, westra2011extending}.

\section{Conclusion}
As people express their thoughts on social media with ease, we observe an opportunity to simulate human-in-the-loop triaging for by examining the interpersonal risk factors (IRF) in human writing for early detection of depression. In this research, we introduce Interpretable Prompting (InterPrompt) as a method to fine-tune the GPT-3 model, resulting in the generation of responses that are both coherent and contextually relevant. We achieve this by combining the multi-task prompting in story completion, enabling it to learn higher-level ways of manipulating the language. As such, we emphasise the importance of combining the \textit{text labels} and the \textit{explanations} that aid the system-level explanability with a special focus on comprehending textual cues projecting reasons behind the presence of \textsc{TBe} and \textsc{PBu}.
Our model outperforms all the baselines both as a \textit{classifier} and \textit{explanation generator}. %The interplay of textual cues in a given text that represents IRF is effectively captured using the InterPrompt approach. 
In the near future, it will be interesting to determine the extent of associations among both IRFs in the near future. Also, our task-specific fine-tuning opens up novel avenues for conducting fine-grained analysis of mental health.%, suggesting Babbage and Curie as the best models %(F-score: 12, Accuracy: ) and Curie () over MentalBERT 
% for TBe and PBu, respectively. %with XXX\% of accuracy and XXX\% of F-score, and in terms of explanation generation with ROUGE-1 score: . 
 %In future, we plan to design and develop more comprehensive, explainable, and safe models by infusing knowledge through clinical questionnaires. 

%\bibliographystyle{ACM-Reference-Format}
%\bibliography{custom}

\begin{thebibliography}{25}

%%% the Web address.

\ifx \showCODEN    \undefined \def \showCODEN     #1{\unskip}     \fi
\ifx \showDOI      \undefined \def \showDOI       #1{#1}\fi
\ifx \showISBNx    \undefined \def \showISBNx     #1{\unskip}     \fi
\ifx \showISBNxiii \undefined \def \showISBNxiii  #1{\unskip}     \fi
\ifx \showISSN     \undefined \def \showISSN      #1{\unskip}     \fi
\ifx \showLCCN     \undefined \def \showLCCN      #1{\unskip}     \fi
\ifx \shownote     \undefined \def \shownote      #1{#1}          \fi
\ifx \showarticletitle \undefined \def \showarticletitle #1{#1}   \fi
\ifx \showURL      \undefined \def \showURL       {\relax}        \fi
% invisible to TeX
\providecommand\bibfield[2]{#2}
\providecommand\bibinfo[2]{#2}
\providecommand\natexlab[1]{#1}
\providecommand\showeprint[2][]{arXiv:#2}

\bibitem[Brown et~al\mbox{.}(2020)]%
        {brown2020language}
\bibfield{author}{\bibinfo{person}{Tom Brown}, \bibinfo{person}{Benjamin Mann}, \bibinfo{person}{Nick Ryder}, \bibinfo{person}{Melanie Subbiah}, \bibinfo{person}{Jared~D Kaplan}, \bibinfo{person}{Prafulla Dhariwal}, \bibinfo{person}{Arvind Neelakantan}, \bibinfo{person}{Pranav Shyam}, \bibinfo{person}{Girish Sastry}, \bibinfo{person}{Amanda Askell}, {et~al\mbox{.}}} \bibinfo{year}{2020}\natexlab{}.
\newblock \showarticletitle{Language models are few-shot learners}.
\newblock \bibinfo{journal}{\emph{Advances in neural information processing systems}}  \bibinfo{volume}{33} (\bibinfo{year}{2020}), \bibinfo{pages}{1877--1901}.
\newblock


\bibitem[Daws(2020)]%
        {artificialintelligencenewsBabylonHealth}
\bibfield{author}{\bibinfo{person}{Ryan Daws}.} \bibinfo{year}{2020}\natexlab{}.
\newblock \showarticletitle{{B}abylon {H}ealth lashes out at doctor who raised {A}{I} chatbot safety concerns}.
\newblock \bibinfo{journal}{\emph{\url{https://www.artificialintelligence-news.com/2020/02/26/babylon-health-doctor-ai-chatbot-safety-concerns/}}} (\bibinfo{year}{2020}).
\newblock


\bibitem[Devlin et~al\mbox{.}(2018)]%
        {devlin2018bert}
\bibfield{author}{\bibinfo{person}{Jacob Devlin}, \bibinfo{person}{Ming-Wei Chang}, \bibinfo{person}{Kenton Lee}, {and} \bibinfo{person}{Kristina Toutanova}.} \bibinfo{year}{2018}\natexlab{}.
\newblock \showarticletitle{Bert: Pre-training of deep bidirectional transformers for language understanding}.
\newblock \bibinfo{journal}{\emph{arXiv preprint arXiv:1810.04805}} (\bibinfo{year}{2018}).
\newblock


\bibitem[Garg et~al\mbox{.}(2023)]%
        {garg2023annotated}
\bibfield{author}{\bibinfo{person}{Muskan Garg}, \bibinfo{person}{Amirmohammad Shahbandegan}, \bibinfo{person}{Amrit Chadha}, {and} \bibinfo{person}{Vijay Mago}.} \bibinfo{year}{2023}\natexlab{}.
\newblock \showarticletitle{An Annotated Dataset for Explainable Interpersonal Risk Factors of Mental Disturbance in Social Media Posts}.
\newblock \bibinfo{journal}{\emph{arXiv preprint arXiv:2305.18727}} (\bibinfo{year}{2023}).
\newblock


\bibitem[Garreau and von Luxburg(2020)]%
        {lime}
\bibfield{author}{\bibinfo{person}{Damien Garreau} {and} \bibinfo{person}{Ulrike von Luxburg}.} \bibinfo{year}{2020}\natexlab{}.
\newblock \bibinfo{title}{Explaining the Explainer: A First Theoretical Analysis of LIME}.
\newblock
\newblock
\showeprint[arxiv]{2001.03447}~[cs.LG]


\bibitem[Hastie et~al\mbox{.}(2009)]%
        {p-value}
\bibfield{author}{\bibinfo{person}{Trevor Hastie}, \bibinfo{person}{Robert Tibshirani}, \bibinfo{person}{Jerome~H Friedman}, {and} \bibinfo{person}{Jerome~H Friedman}.} \bibinfo{year}{2009}\natexlab{}.
\newblock \bibinfo{booktitle}{\emph{The elements of statistical learning: data mining, inference, and prediction}}. Vol.~\bibinfo{volume}{2}.
\newblock \bibinfo{publisher}{Springer}.
\newblock


\bibitem[James et~al\mbox{.}(2013)]%
        {t-test}
\bibfield{author}{\bibinfo{person}{Gareth James}, \bibinfo{person}{Daniela Witten}, \bibinfo{person}{Trevor Hastie}, {and} \bibinfo{person}{Robert Tibshirani}.} \bibinfo{year}{2013}\natexlab{}.
\newblock \bibinfo{booktitle}{\emph{An introduction to statistical learning}}. Vol.~\bibinfo{volume}{112}.
\newblock \bibinfo{publisher}{Springer}.
\newblock


\bibitem[Ji et~al\mbox{.}(2021)]%
        {ji2021mentalbert}
\bibfield{author}{\bibinfo{person}{Shaoxiong Ji}, \bibinfo{person}{Tianlin Zhang}, \bibinfo{person}{Luna Ansari}, \bibinfo{person}{Jie Fu}, \bibinfo{person}{Prayag Tiwari}, {and} \bibinfo{person}{Erik Cambria}.} \bibinfo{year}{2021}\natexlab{}.
\newblock \showarticletitle{Mentalbert: Publicly available pretrained language models for mental healthcare}.
\newblock \bibinfo{journal}{\emph{arXiv preprint arXiv:2110.15621}} (\bibinfo{year}{2021}).
\newblock


\bibitem[Joiner~Jr et~al\mbox{.}(2009)]%
        {joiner2009interpersonal}
\bibfield{author}{\bibinfo{person}{Thomas~E Joiner~Jr}, \bibinfo{person}{Kimberly~A Van~Orden}, \bibinfo{person}{Tracy~K Witte}, {and} \bibinfo{person}{M~David Rudd}.} \bibinfo{year}{2009}\natexlab{}.
\newblock \bibinfo{booktitle}{\emph{The interpersonal theory of suicide: Guidance for working with suicidal clients.}}
\newblock \bibinfo{publisher}{American Psychological Association}.
\newblock


\bibitem[Kalpi{\'{c}} et~al\mbox{.}(2011)]%
        {t-test2}
\bibfield{author}{\bibinfo{person}{Damir Kalpi{\'{c}}}, \bibinfo{person}{Nikica Hlupi{\'{c}}}, {and} \bibinfo{person}{Miodrag Lovri{\'{c}}}.} \bibinfo{year}{2011}\natexlab{}.
\newblock \bibinfo{booktitle}{\emph{Student's t-Tests}}.
\newblock \bibinfo{publisher}{Springer Berlin Heidelberg}, \bibinfo{address}{Berlin, Heidelberg}, \bibinfo{pages}{1559--1563}.
\newblock
\showISBNx{978-3-642-04898-2}
\urldef\tempurl%
\url{https://doi.org/10.1007/978-3-642-04898-2_641}
\showDOI{\tempurl}


\bibitem[Lin(2004)]%
        {lin2004rouge}
\bibfield{author}{\bibinfo{person}{Chin-Yew Lin}.} \bibinfo{year}{2004}\natexlab{}.
\newblock \showarticletitle{Rouge: A package for automatic evaluation of summaries}. In \bibinfo{booktitle}{\emph{Text summarization branches out}}. \bibinfo{pages}{74--81}.
\newblock


\bibitem[Liu et~al\mbox{.}(2019)]%
        {liu2019roberta}
\bibfield{author}{\bibinfo{person}{Yinhan Liu}, \bibinfo{person}{Myle Ott}, \bibinfo{person}{Naman Goyal}, \bibinfo{person}{Jingfei Du}, \bibinfo{person}{Mandar Joshi}, \bibinfo{person}{Danqi Chen}, \bibinfo{person}{Omer Levy}, \bibinfo{person}{Mike Lewis}, \bibinfo{person}{Luke Zettlemoyer}, {and} \bibinfo{person}{Veselin Stoyanov}.} \bibinfo{year}{2019}\natexlab{}.
\newblock \showarticletitle{Roberta: A robustly optimized bert pretraining approach}.
\newblock \bibinfo{journal}{\emph{arXiv preprint arXiv:1907.11692}} (\bibinfo{year}{2019}).
\newblock


\bibitem[Lundberg and Lee(2017)]%
        {shap}
\bibfield{author}{\bibinfo{person}{Scott Lundberg} {and} \bibinfo{person}{Su-In Lee}.} \bibinfo{year}{2017}\natexlab{}.
\newblock \bibinfo{title}{A Unified Approach to Interpreting Model Predictions}.
\newblock
\newblock
\showeprint[arxiv]{1705.07874}~[cs.AI]


\bibitem[Papineni et~al\mbox{.}(2002)]%
        {papineni2002bleu}
\bibfield{author}{\bibinfo{person}{Kishore Papineni}, \bibinfo{person}{Salim Roukos}, \bibinfo{person}{Todd Ward}, {and} \bibinfo{person}{Wei-Jing Zhu}.} \bibinfo{year}{2002}\natexlab{}.
\newblock \showarticletitle{Bleu: a method for automatic evaluation of machine translation}. In \bibinfo{booktitle}{\emph{Proceedings of the 40th annual meeting of the Association for Computational Linguistics}}. \bibinfo{pages}{311--318}.
\newblock


\bibitem[Roy et~al\mbox{.}(2023)]%
        {roy2023proknow}
\bibfield{author}{\bibinfo{person}{Kaushik Roy}, \bibinfo{person}{Manas Gaur}, \bibinfo{person}{Misagh Soltani}, \bibinfo{person}{Vipula Rawte}, \bibinfo{person}{Ashwin Kalyan}, {and} \bibinfo{person}{Amit Sheth}.} \bibinfo{year}{2023}\natexlab{}.
\newblock \showarticletitle{ProKnow: Process knowledge for safety constrained and explainable question generation for mental health diagnostic assistance}.
\newblock \bibinfo{journal}{\emph{Frontiers in big Data}}  \bibinfo{volume}{5} (\bibinfo{year}{2023}), \bibinfo{pages}{1056728}.
\newblock


\bibitem[Sanh et~al\mbox{.}(2019)]%
        {sanh2019distilbert}
\bibfield{author}{\bibinfo{person}{Victor Sanh}, \bibinfo{person}{Lysandre Debut}, \bibinfo{person}{Julien Chaumond}, {and} \bibinfo{person}{Thomas Wolf}.} \bibinfo{year}{2019}\natexlab{}.
\newblock \showarticletitle{DistilBERT, a distilled version of BERT: smaller, faster, cheaper and lighter}.
\newblock \bibinfo{journal}{\emph{arXiv preprint arXiv:1910.01108}} (\bibinfo{year}{2019}).
\newblock


\bibitem[Sawhney et~al\mbox{.}(2023)]%
        {sawhney2023much}
\bibfield{author}{\bibinfo{person}{Ramit Sawhney}, \bibinfo{person}{Atula Neerkaje}, \bibinfo{person}{Ivan Habernal}, {and} \bibinfo{person}{Lucie Flek}.} \bibinfo{year}{2023}\natexlab{}.
\newblock \showarticletitle{How Much User Context Do We Need? Privacy by Design in Mental Health NLP Applications}. In \bibinfo{booktitle}{\emph{Proceedings of the International AAAI Conference on Web and Social Media}}, Vol.~\bibinfo{volume}{17}. \bibinfo{pages}{766--776}.
\newblock


\bibitem[Tay et~al\mbox{.}(2018)]%
        {tay2018mental}
\bibfield{author}{\bibinfo{person}{Stacie Tay}, \bibinfo{person}{Kat Alcock}, {and} \bibinfo{person}{Katrina Scior}.} \bibinfo{year}{2018}\natexlab{}.
\newblock \showarticletitle{Mental health problems among clinical psychologists: Stigma and its impact on disclosure and help-seeking}.
\newblock \bibinfo{journal}{\emph{Journal of Clinical Psychology}} \bibinfo{volume}{74}, \bibinfo{number}{9} (\bibinfo{year}{2018}), \bibinfo{pages}{1545--1555}.
\newblock


\bibitem[Torous et~al\mbox{.}(2021)]%
        {torous2021growing}
\bibfield{author}{\bibinfo{person}{John Torous}, \bibinfo{person}{Sandra Bucci}, \bibinfo{person}{Imogen~H Bell}, \bibinfo{person}{Lars~V Kessing}, \bibinfo{person}{Maria Faurholt-Jepsen}, \bibinfo{person}{Pauline Whelan}, \bibinfo{person}{Andre~F Carvalho}, \bibinfo{person}{Matcheri Keshavan}, \bibinfo{person}{Jake Linardon}, {and} \bibinfo{person}{Joseph Firth}.} \bibinfo{year}{2021}\natexlab{}.
\newblock \showarticletitle{The growing field of digital psychiatry: current evidence and the future of apps, social media, chatbots, and virtual reality}.
\newblock \bibinfo{journal}{\emph{World Psychiatry}} \bibinfo{volume}{20}, \bibinfo{number}{3} (\bibinfo{year}{2021}), \bibinfo{pages}{318--335}.
\newblock


\bibitem[Wei et~al\mbox{.}({[n.\,d.]})]%
        {weifinetuned}
\bibfield{author}{\bibinfo{person}{Jason Wei}, \bibinfo{person}{Maarten Bosma}, \bibinfo{person}{Vincent Zhao}, \bibinfo{person}{Kelvin Guu}, \bibinfo{person}{Adams~Wei Yu}, \bibinfo{person}{Brian Lester}, \bibinfo{person}{Nan Du}, \bibinfo{person}{Andrew~M Dai}, {and} \bibinfo{person}{Quoc~V Le}.} \bibinfo{year}{[n.\,d.]}\natexlab{}.
\newblock \showarticletitle{Finetuned Language Models are Zero-Shot Learners}. In \bibinfo{booktitle}{\emph{International Conference on Learning Representations}}.
\newblock


\bibitem[Westra et~al\mbox{.}(2011)]%
        {westra2011extending}
\bibfield{author}{\bibinfo{person}{Henny~A Westra}, \bibinfo{person}{Adi Aviram}, {and} \bibinfo{person}{Faye~K Doell}.} \bibinfo{year}{2011}\natexlab{}.
\newblock \showarticletitle{Extending motivational interviewing to the treatment of major mental health problems: current directions and evidence}.
\newblock \bibinfo{journal}{\emph{The Canadian Journal of Psychiatry}} (\bibinfo{year}{2011}).
\newblock


\bibitem[Xie et~al\mbox{.}(2022)]%
        {xie2022pre}
\bibfield{author}{\bibinfo{person}{Qianqian Xie}, \bibinfo{person}{Jennifer~Amy Bishop}, \bibinfo{person}{Prayag Tiwari}, {and} \bibinfo{person}{Sophia Ananiadou}.} \bibinfo{year}{2022}\natexlab{}.
\newblock \showarticletitle{Pre-trained language models with domain knowledge for biomedical extractive summarization}.
\newblock \bibinfo{journal}{\emph{Knowledge-Based Systems}}  \bibinfo{volume}{252} (\bibinfo{year}{2022}), \bibinfo{pages}{109460}.
\newblock


\bibitem[Yang et~al\mbox{.}(2023)]%
        {yang2023evaluations}
\bibfield{author}{\bibinfo{person}{Kailai Yang}, \bibinfo{person}{Shaoxiong Ji}, \bibinfo{person}{Tianlin Zhang}, \bibinfo{person}{Qianqian Xie}, {and} \bibinfo{person}{Sophia Ananiadou}.} \bibinfo{year}{2023}\natexlab{}.
\newblock \showarticletitle{On the Evaluations of ChatGPT and Emotion-enhanced Prompting for Mental Health Analysis}.
\newblock \bibinfo{journal}{\emph{arXiv preprint arXiv:2304.03347}} (\bibinfo{year}{2023}).
\newblock


\bibitem[Yoo et~al\mbox{.}(2023)]%
        {yoo2023discussing}
\bibfield{author}{\bibinfo{person}{Dong~Whi Yoo}, \bibinfo{person}{Aditi Bhatnagar}, \bibinfo{person}{Sindhu~Kiranmai Ernala}, \bibinfo{person}{Asra Ali}, \bibinfo{person}{Michael~L Birnbaum}, \bibinfo{person}{Gregory~D Abowd}, {and} \bibinfo{person}{Munmun De~Choudhury}.} \bibinfo{year}{2023}\natexlab{}.
\newblock \showarticletitle{Discussing Social Media During Psychotherapy Consultations: Patient Narratives and Privacy Implications}.
\newblock \bibinfo{journal}{\emph{Proceedings of the ACM on Human-Computer Interaction}} \bibinfo{volume}{7}, \bibinfo{number}{CSCW1} (\bibinfo{year}{2023}), \bibinfo{pages}{1--24}.
\newblock


\bibitem[Zirikly et~al\mbox{.}(2019)]%
        {zirikly2019clpsych}
\bibfield{author}{\bibinfo{person}{Ayah Zirikly}, \bibinfo{person}{Philip Resnik}, \bibinfo{person}{Ozlem Uzuner}, {and} \bibinfo{person}{Kristy Hollingshead}.} \bibinfo{year}{2019}\natexlab{}.
\newblock \showarticletitle{CLPsych 2019 shared task: Predicting the degree of suicide risk in Reddit posts}. In \bibinfo{booktitle}{\emph{Proceedings of the sixth workshop on computational linguistics and clinical psychology}}. \bibinfo{pages}{24--33}.
\newblock


\end{thebibliography}
%\input{output.bbl}

\end{document}